\documentclass[letterpaper, 10 pt, conference]{ieeeconf}
\IEEEoverridecommandlockouts
\usepackage{cite}

\usepackage{graphicx}

\usepackage[bookmarks=true]{hyperref}
\hypersetup{%
  breaklinks,
  bookmarksnumbered=true,
  bookmarksopen=true,
  colorlinks=true,
  linkcolor=black,
  urlcolor=black,
  citecolor=black
}

\usepackage{amsmath}
\usepackage{amsfonts}
\usepackage{comment}
\def \nGamma {\mathnormal{\Gamma}}

\usepackage{array}

\makeatletter
\newcommand{\thickhline}{%
    \noalign {\ifnum 0=`}\fi \hrule height 1.5pt
    \futurelet \reserved@a \@xhline
}
\newcolumntype{"}{@{\hskip\tabcolsep\vrule width 1.5pt\hskip\tabcolsep}}
\makeatother

\begin{document}
\bstctlcite{IEEEexample:BSTcontrol}

\title{\LARGE \bf Hybrid Event Shaping \\ to Stabilize Periodic Hybrid Orbits}

\author{James Zhu, Nathan J. Kong, George Council, and Aaron M. Johnson %
    \thanks{This material is based upon work supported by the National Science Foundation
    under grant \#CMMI-1943900 and \#ECCS-1924723 as well as the U.S. Army Research
    Office under grant \#W911NF-19-1-0080. The views and conclusions contained
    in this document are those of the authors and should not be interpreted as
    representing the official policies, either expressed or implied, of the National Science Foundation,
    the Army Research Office, or the U.S. Government.
    The U.S. Government is authorized to reproduce and distribute reprints
    for Government purposes notwithstanding any copyright notation herein.}%
    \thanks{All authors are with the Department of Mechanical Engineering, Carnegie Mellon University, Pittsburgh, PA, USA, \texttt{jameszhu@andrew.cmu.edu}}%
}

\maketitle
\thispagestyle{empty}
\pagestyle{empty}

%%%%%%%%%%%%%%%%%%%%%%%%%%%%%%%%%%%%%%%%%%%%%%%%%%%%%%%%%%%%%%%%%%%%%%%%%%%%%%%%

\begin{abstract}
    Many controllers for legged robotic systems leverage open- or closed-loop control at discrete hybrid events to enhance stability. 
    These controllers appear in several well studied phenomena such as the Raibert stepping controller, paddle juggling, and swing leg retraction. 
    This work introduces hybrid event shaping (HES): a generalized method for analyzing and designing stable hybrid event controllers. 
    HES utilizes the saltation matrix, which gives a closed-form equation for the effect that hybrid events have on stability.
    We also introduce shape parameters, which are higher order terms that can be tuned completely independently of the system dynamics to promote stability.
    Optimization methods are used to produce values of these parameters that optimize a stability measure.
    Hybrid event shaping captures previously developed control methods while also producing new optimally stable trajectories without the need for continuous-domain feedback.
\end{abstract}

\section{Introduction}
    
    In general, the walking and running gaits of legged robots are naturally unstable and challenging to control. 
    %A solution to prevent robots from failing is to implement feedback control. 
    %Early work by Raibert successfully used a feedback controller to modulate leg touchdown angle and stabilize a one-legged hopping robot \cite{raibert}.
    %More recent work has taken advantage of improved computing technologies to develop high frequency feedback controllers for hybrid systems \cite{rabbit,hzd,Altin_hybrid_MPC}.
    Hybrid systems such as these are difficult to work with due to the discontinuities in state and dynamics that occur at hybrid events, such as toe touchdown.
    These discontinuities violate assumptions of standard controllers designed for purely continuous systems, and work is ongoing to adapt these controllers for hybrid systems \cite{Altin_hybrid_MPC,kong2021ilqr}.
    One strategy for hybrid control is to cancel out the effects of hybrid events by working with an invariant subsystem \cite{invariant_impact_control,Burden_dimension_reduction,dbhop}.
    We propose instead that the effects of hybrid events are valuable due to rich control properties that can be used to stabilize trajectories of a hybrid system.
    
\begin{figure}[t]
        \centering
        \includegraphics[width=1\linewidth]{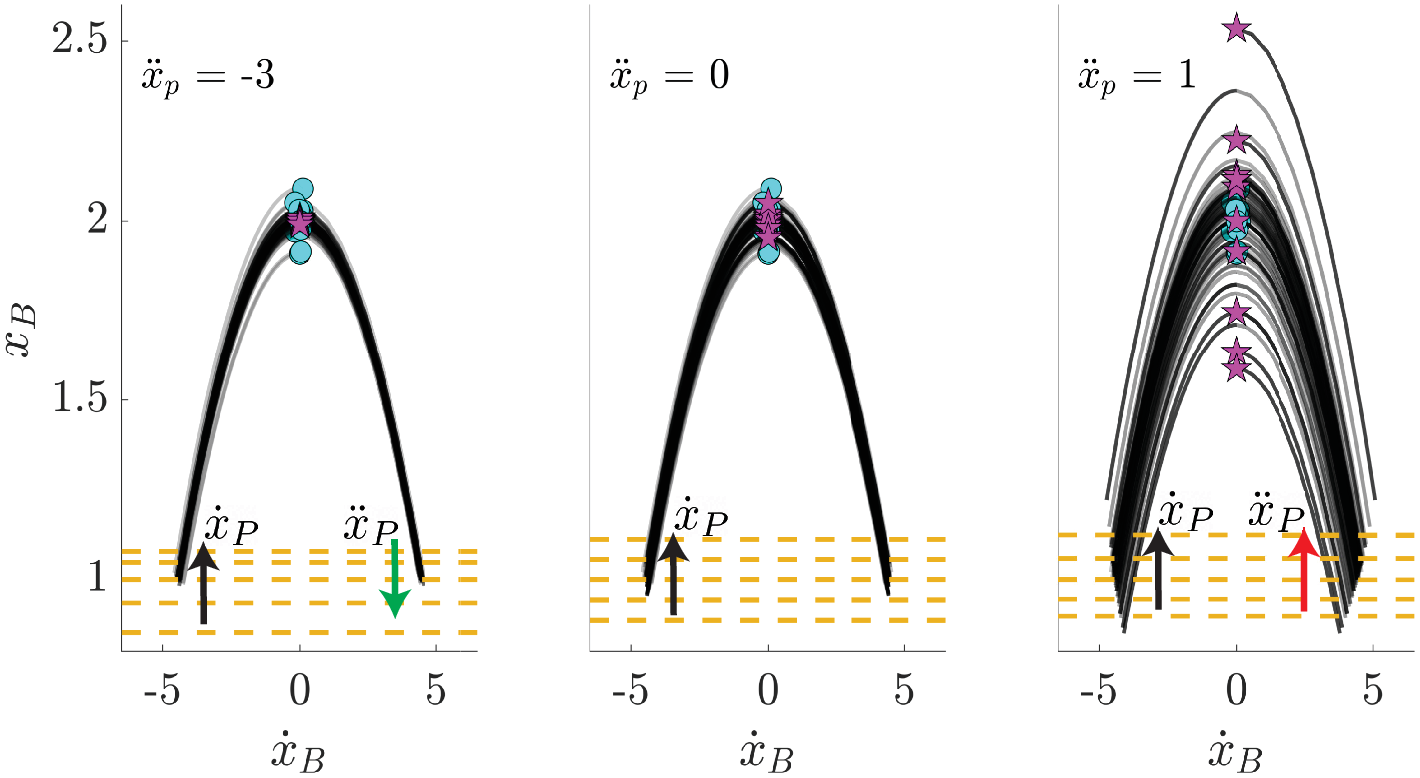}
        \caption{The paddle juggler system \cite{paddlejuggling} has no control authority while the ball is in the air. The paddle acceleration at impact determines the convergence/divergence of the system from initial points (cyan dots) to the final states (magenta stars) after 5 cycles. This example underscores how hybrid event shaping can stabilize a periodic hybrid system.}
        \label{fig:paddle_juggler}
    \end{figure}

    Several works have examined the utility of controlling hybrid event conditions to improve system stability without any closed-loop continuous-domain control \cite{paddlejuggling,swinglegretraction,ankarali2010control}. 
    For example, \cite{paddlejuggling} found that for the paddle juggler system, paddle acceleration at impact uniquely determines the local stability properties of a periodic trajectory, Fig.~\ref{fig:paddle_juggler}. 
    Other works \cite{dbhop,Green_2020} generated open-loop swing leg trajectories that produced deadbeat hopping of a SLIP-like system.
    Each of these works found that controlling a hybrid system only at the moment of a hybrid event is sufficient to provide stabilization.
    So far, however, these results have only been produced for each specific problem structure and a clear connection between these works has yet to be established.
    
    In this work, we propose the concept of hybrid event shaping (HES), which describes how hybrid event parameters can be chosen to affect the stability properties of a periodic orbit.
    We also propose methods to produce values of these hybrid event parameters to optimize a stability measure of a trajectory.
    This approach is tested on both existing examples from \cite{paddlejuggling,swinglegretraction} and on a new bipedal robot controller.
    
\begin{comment}
    This paper is organized as follows.
    Section II formally defines hybrid systems and introduces the saltation matrix, which will be used analyze stability of periodic hybrid systems. 
    Section III introduces the novel work of hybrid event shaping and demonstrates how the discrete nature of hybrid events can be leveraged to improve stability. 
    Section IV presents examples of HES methods stabilizing naturally unstable hybrid systems
    and Section V concludes with a discussion on the implications and future work.
\end{comment}
    
\section{Preliminaries}

    \subsection{Hybrid systems}
    
%    \begin{itemize}
%        \item Define hybrid systems
%    \end{itemize}
    
    Hybrid systems are a class of dynamical systems which exhibit both continuous and discrete dynamics \cite{vanderschaft_hybrid_systems,lygeros_hybrid_automata}. 
    Following the adaptation of \cite{johnson_hybrid_systems} in \cite{kong2021ilqr}, we define a $C^r$ hybrid dynamical system for continuity class $r\in \mathbb{N}_{>0} \cup \{\infty,\omega\}$ as a tuple $\mathcal{H}:=(\mathcal{J},\nGamma,\mathcal{D},\mathcal{F},\mathcal{G},\mathcal{R})$ where:
    
    \begin{enumerate}
        \item $\mathcal{J} := \{I,J,\hdots,K\}\subset\mathbb{N}$ is the set of discrete modes.
        \item $\nGamma\subset\mathcal{J}\times\mathcal{J}$ is the set of discrete transitions forming a directed graph structure over $\mathcal{J}$.
        \item $\mathcal{D} := \amalg_{I\in\mathcal{J}}D_I$ is the collection of domains where $D_I$ is a $C^r$ manifold with corners \cite{Joyce2012}.
        \item $\mathcal{F} := \amalg_{I\in\mathcal{J}}F_I$ is a collection of time-varying vector fields with control inputs $F_I:\mathbb{R}\times D_I\times U_I\rightarrow TD_I$ where $U_I$ is the space of allowable control inputs.
        \item $\mathcal{G} := \amalg_{(I,J)\in\nGamma}G_{(I,J)}(t)$ is a collection of guards where $G_{(I,J)}(t)\subset D_I \times U_I$ for each $(I,J)\in\nGamma$ is defined as a sublevel set of a $C^r$ function, i.e.\ $G_{(I,J)}(t)=\{(x,u)\in D_I\times U_I|g_{(I,J)}(t,x,u)\leq 0\}$.
        \item $\mathcal{R}:\mathbb{R}\times\mathcal{G}\rightarrow\mathcal{D}$ is a $C^r$ map called the reset map that restricts as $R_{(I,J)} := \mathcal{R}|_{G_{(I,J)}(t)}:G_{(I,J)}(t)\rightarrow D_J$ for each $(I,J)\in\nGamma$.
    \end{enumerate}
    
    An execution of a hybrid system \cite{johnson_hybrid_systems} begins at an initial state $x_0\in D_I$. 
    With a particular input $u_I(t,x)$, the system follows the dynamics $F_I$ on $D_I$. 
    If the system reaches the guard surface $G_{(I,J)}$, the reset map $R_{(I,J)}$ is applied and the system continues in domain $D_J$ with the corresponding dynamics defined by $F_J$. 
    An execution of a hybrid system is defined over a hybrid time domain which is a disjoint union of intervals $\amalg_{j\in\mathcal{N}}[\underbar{t}_j,\bar{t}_j]$. 
    The flow $\phi(t,t_0,x_0)$ describes how the system evolves from some initial time $t_0$ and state $x_0$ for some length of execution time $t$.
    
    Hybrid systems may exhibit complex behaviors including sliding \cite{sliding}, branching \cite{simic_towards_2000}, and Zeno phenomena where infinite transitions occur in finite time \cite{zeno}. 
    Following prior literature \cite{aizerman_hybrid_stability,contraction,Leine}, we assume these behaviors do not occur, such that guard surfaces are isolated and intersected transversely \cite{lygeros_hybrid_automata, johnson_hybrid_systems} and no Zeno executions occur. 
    These assumptions are not generally detrimental to the validity of this theory to applications like legged locomotion.
    
    \subsection{Saltation matrix}
    
%       \begin{itemize}
%        \item Saltation matrices map executions  perturbations across hybrid events
%        \item Dependent on flows, guards, and resets (and derivatives)
%        \item From the derivatives, shape parameter terms may exist in saltation matrix independent of continuous flow
%    \end{itemize}
    
     Hybrid systems can be divided into continuous domains and discrete hybrid events. 
     For each of these components, the linearized variational equations \cite{hirsch2012differential} describe how perturbations of state at the beginning of the phase evolve to the end of the phase.
     %These variational equations are necessary to analyze the stability of the full periodic orbit. 
     In continuous domains, variational equations can be derived and discretized into a mapping $A_I(\bar{t}_j-\underline{t}_j,\underline{t}_j,x(\underline{t}_j))$ equivalent to the linearized discrete dynamics matrix in $x_{k+1}\approx A_Ix_k+B_Iu$ \cite{hirsch2012differential}.
     The variational equations of hybrid events are characterized by the saltation matrix $\Xi_{(I,J)}(\bar{t}_i,x(\bar{t}_i),u(\bar{t}_i))$, which describes the transition between modes $I$ and $J$ that occurs at time $\bar{t}_i$ with $x(\bar{t}_i) \in G_{(I,J)}$ and some input $u(\bar{t}_i)$.
     The saltation matrix approximates the first-order change in perturbations in state before the hybrid event at $\delta x(\bar{t}_i)$ to perturbations afterward $\delta x(\underline{t}_{i+1})$ \cite{contraction} such that:
    \begin{equation}
        \delta x(\underbar{t}_{i+1}) = \Xi_{(I,J)}(\bar{t}_i,x(\bar{t}_i),u(\bar{t}_i))\delta x(\bar{t}_i)+\text{h.o.t.}
    \end{equation}
    where $\text{h.o.t.}$ represents higher order terms.
    Following the formulation from \cite{contraction,Muller_lyapunov_exponents}, the saltation matrix is,
    \begin{equation}
        \Xi=D_xR+\frac{(F_J-D_xR\cdot F_I-
        D_tR)D_xg}{D_tg+D_xgF_I}
        \label{eq:salt}
    \end{equation}
    where
    \begin{align*}
        \Xi &:= \Xi_{(I,J)}(\bar{t}_i,x(\bar{t}_i),u(\bar{t}_i)), \;R := R_{(I,J)}(\bar{t}_i,x(\bar{t}_i),u(\bar{t}_i)), \\ 
        g  &:= g_{(I,J)}(\bar{t}_i,x(\bar{t}_i),u(\bar{t}_i)), \;
        F_I := F_I(\bar{t}_i,x(\bar{t}_i),u(\bar{t}_i)), \\ 
        F_J &:= F_J(\underline{t}_{i+1},R_{(I,J)}(\bar{t}_i,x(\bar{t}_i),u(\bar{t}_i))
    \end{align*}
    
    \subsection{Periodic Stability Analysis}
    
%        \begin{itemize}
%            \item Burden looked at contraction conditions for saltation matrices
%            \item Contraction is not feasible in many cases, looking at stability is much more reasonable
%            \item Local stability tells us nothing in a hybrid system
%            \item We must combine variational equations in continuous domains and hybrid events to understand global                 stability 
%            \item Spectral radius of monodromy matrix is stability measure
%            \item Measure can not be below 1 for autonomous systems, not true for non-autonomous systems
%            \item Even without controlling the continuous dynamics, shaping the saltation matrices alone can induce                 stability from typically unstable systems
%            \item We can also look at norms if we want
%        \end{itemize}
        
        A dynamical system has a periodic trajectory (orbit) with period $T$ if for some initial condition $x_0$, there exists a solution $\phi(t, t_0, x_0)$ where $\phi(t, t_0, x_0) = \phi(t+kT, t_0, x_0)$ for all $t$ and $k$ \cite{Leine}. 
        Perturbations $\delta x_0$ around $x_0$ can be mapped to perturbations $\delta x_T$ after period $T$ by a linearized mapping known as the monodromy matrix, $\Phi$ \cite{monodromy}: 
        \begin{equation}
            \Phi = \frac{\partial\phi(T+t_0,t_0,x_0)}{\partial x_0}
        \end{equation}
        such that to the first order, $\delta x_T \approx \Phi\delta x_0$.
        
        Following a formulation similar to \cite{monodromy_matrix}, the monodromy matrix can be computed by sequentially composing  the linearized variational equations in each continuous domain ($A$) and the saltation matrices ($\Xi$) at each hybrid event. 
        For a hybrid periodic orbit with $N$ domains, the monodromy matrix can be formulated as:
        \begin{align}
            \Phi = \Xi_{(N,1)}A_N\dots \Xi_{(2,3)}A_2\Xi_{(1,2)}A_1
        \end{align}
        
        The monodromy matrix determines local asymptotic orbital stability (which we refer to simply as stability).
        For nonautonomous systems, stability is determined by the maximum magnitude of the eigenvalues, $\max(|\lambda|)$ \cite{Leine}.
        We refer to this as the stability measure, $\psi$, where a trajectory is stable when $\psi<1$.
        Autonomous systems always have an eigenvalue that is equal to 1 since for non-time varying dynamics, perturbations along the flow of the orbit will by definition map back to themselves after period $T$ \cite{Leine}.
        Assuming non-convergence in this direction is allowable, $\psi$ for autonomous systems is based on the remaining eigenvalues.

\section{Methods}

%    \begin{itemize}
%        \item The stability of an orbit changes continuously with changes in the timing and state of hybrid events, as well as so called shape parameters
%        \item These shape parameters are particularly special because they can tune stability independently of a nominal orbit
%        \item By changing the guard trajectory around (but not exactly at) a nominal impact point, the stability                   properties of a given orbit can changed without changing the orbit itself
%    \end{itemize}

    \subsection{Hybrid event shaping}
    
%    \begin{itemize}
%        \item In general, saltation matrices may be a function of time and state, but also potentially higher order                ``shape parameters''
%        \item Many simplified systems have guards (and/or reset maps) that can be controlled independently from the                continuous flow.
%        \item By changing the guard trajectory around (but not exactly at) a nominal impact point, the stability                   properties of a given orbit can changed without changing the orbit itself
%        \item Present paddle juggler case
%    \end{itemize}

    %The primary utility of the saltation matrix in this setting is the ability to explicitly reason about how hybrid events can affect the shape of perturbations. 
    
    Hybrid events can greatly affect the stability of an orbit due to the unbounded discontinuous changes that are made to perturbations.
    The saltation matrix allows for an explicit understanding of how to perform ``hybrid event shaping'' (HES), i.e.\ choosing hybrid event parameters (such as timing, state, input, and higher order ``shape parameters'') to improve the stability of a periodic trajectory.
    The key insight is that hybrid event shaping introduces a generalizable method to stabilize hybrid systems that is independent of continuous-domain control, but that can work in concert with it.

    In general, the open-loop continuous variational equations of a hybrid system are functions of initial and final time, initial state, and system dynamics. 
    However, it is challenging to alter any of these parameters because changes will propagate through the rest of the trajectory and periodicity may be violated, though we present a trajectory optimization method below to handle this. 
    The saltation matrix is a function of nominal event time, state and dynamics, but additionally may be a function of higher order shape parameters $h$ that do not influence the dynamics of the system.
    These parameters arise from the derivatives of the guards ($D_xg$ and $D_tg$) and reset maps ($D_xR$ and $D_tR$) but are not present in the guard, reset map, or vector field definitions themselves.
    Therefore, shape parameters have absolutely no effect on the nominal trajectory and can be chosen completely freely.
    
    One example of a shape parameter is the angular velocity of a massless leg of a spring-loaded inverted pendulum.
    Since a massless leg does not induce any torque in the air or forces at touchdown, only the position of the leg at touchdown affects the trajectory of the body.
    However, leg velocity appears in the saltation matrix and has a significant effect on orbital stability \cite{swinglegretraction}.
    %Shape variables are also similar to a feedback gain matrix $K$, which has no effect on the nominal trajectory, but is important in stabilizing perturbed states.
    
    For more complex models of robots, there may not be any physical shape parameters that can be tuned.
    For example, legged robots with non-massless legs can not vary leg velocity at impact without also changing its trajectory. These cases can be handled by running a trajectory optimization at the same time as applying HES, as we show in Sec.~\ref{sec:trajopt}, or by adding additional virtual hybrid events.
        
    \subsection{Virtual hybrid events}

%    \begin{itemize}
%        \item A discrete change in input can act as an artificial hybrid event
%        \item This may be useful in adding better ability to stabilize a system
%    \end{itemize}

    Certain control systems naturally have discontinuities in control inputs, such as bang-bang control, sliding mode control, or systems with actuators that have discretized (i.e.\ on-off) inputs.
    These discontinuities in control can cause an instantaneous change in the dynamics of the system, resulting in a virtual (as opposed to physical) hybrid event. 
    %This virtual hybrid event can be parameterized by a choice of guard function, similar to a phase variable studied in biomechanics \cite{phasevariable}, and will have an identity reset map. 
    Virtual hybrid events act no differently than physical hybrid events and induce saltation matrices with shape parameters to be tuned for stability.
    Even for systems where discontinuous control inputs are not necessary, the addition of virtual saltation matrices and shape parameters allows for a greater authority to improve stability.

    \subsection{Stability measure derivative}
    
    Our goal is to determine the optimal choice of hybrid event parameters that minimizes the stability measure of a trajectory.
    Since directly computing eigenvalues in closed-form is not generally feasible, one solution is to use numerical methods to perform optimization \cite{optimal_swing_leg_retraction}.
    However, this strategy becomes untenable for high dimensional problems.
    Instead, by using the saltation matrix formulation \eqref{eq:salt}, derivatives of the stability measure can be directly computed, allowing for the use of more efficient optimization methods and making the problem much more tractable.
    
    Assuming that the monodromy matrix $\Phi$ depends continuously on each shape parameter $h_n$, the eigenvalues of $\Phi$ are always continuous with respect to $h_n$ \cite{IntrotoNumericalAnalysis}.
    For a diagonalizable $\Phi$, the derivative of the eigenvalues with respect to $h_n$ can be computed in closed form \cite{eig_derivatives}. 
    Assume that matrix $\Phi(h_n)$ has simple (non-repeating) eigenvalues, $\lambda_1,\hdots,\lambda_N$, and let $\boldsymbol{j}_i$ and $\boldsymbol{k}_i$ denote the left and right eigenvectors associated with $\lambda_i$. 
    Then the derivative $\frac{d\lambda_i}{dh_n}$ is:
    \begin{align}
        \frac{d\lambda_i}{dh_n}=\boldsymbol{k}_i'\frac{d\Phi}{dh_n}\boldsymbol{j}_i
        \label{eq:eig_derv}
    \end{align}
    For matrices with eigenvalues that repeat, the derivatives of the repeated eigenvalues can be calculated similarly with a matrix of associated eigenvectors \cite{eig_derivatives}.
    
    $\frac{d\Phi}{dh_n}$ can be found using the derivative product rule, which simplifies if each shape parameter only appears in one saltation matrix.
    We make this assumption here to improve computational efficiency, but it is not required generically.
    Without loss of generality, take $\frac{d\Xi_{(1,2)}}{dh_n}\neq 0$, so that:
    \begin{align}
        \frac{d\Phi}{dh_n}=\Xi_{(N,1)}A_N\dots A_{2}\frac{d\Xi_{(1,2)}}{dh_n}A_1
        \label{eq:mono_derv}
    \end{align}
        
    Substituting \eqref{eq:mono_derv} into \eqref{eq:eig_derv} allows us to compute the derivative of the stability measure with respect to each of the shape parameters.
    Eq.~\eqref{eq:eig_derv} is not valid for non-diagonalizable monodromy matrices.
    However, the guaranteed continuity of the stability measure allows for finite-difference methods to be used in any non-diagonalizable cases. 
    
    The derivative computation from \eqref{eq:mono_derv} can be adapted for changes in $x$ and $t$ as well.
    Without loss of generality, consider again $\Xi_{(1,2)}$.
    For hybrid event time $t_{(1,2)}$, the derivative $\frac{d\Xi_{(1,2)}}{dt_{(1,2)}}$ can be computed in closed-form.
    Additionally, the derivatives $\frac{dA_1}{dt_{(1,2)}}$ and $\frac{dA_2}{dt_{(1,2)}}$ are non-zero and can be computed through standard methods \cite{Khalil}.
    The product rule expansion of $\frac{d\Phi}{dt_{(1,2)}}$ consists of additional terms compared to \eqref{eq:mono_derv} but otherwise can be computed similarly. 
    $\frac{d\Phi}{dx_{(1,2)}}$ for hybrid event state $x_{(1,2)}$ can be computed this same way.
    
    \subsection{Shape parameter stability optimization}
    
%    \begin{itemize}
%        \item Given a system with an independently controlled guard, how can we find the optimal shape parameter inputs?
%        \item Best be done with numerical optimization
%        \item We want to minimize spectral radius of monodromy matrix
%        \item The derivatives of this cost function can be computed almost everywhere
%    \end{itemize}

    Optimization techniques \cite{num_opt} are able to select optimal hybrid event parameters that minimize the stability measure.
    Two optimization methods are presented here: the first optimizing the shape parameters without affecting the dynamics of the nominal orbit, and the second optimizing both the hybrid events and periodic orbit simultaneously.

    Shape parameters are powerful because they do not appear in the dynamics of the system and have no effect on the nominal trajectory.
    This means that for a given periodic trajectory, the shape parameters can be chosen freely.
    We use an optimization framework to choose these shape parameters with the goal of optimizing the stability measure $\psi(\Phi(h))$ of a trajectory,
    \begin{equation}
        \begin{array}{cl}\underset{h_{1:M}}{\operatorname{minimize}} & \psi(\Phi(h)) \end{array}
        \label{eq:shape_opt}
    \end{equation}
    
    The ability to compute derivatives of the stability measure allows for this optimization to be more computationally efficient.
    The examples below show how this optimization method is able to reproduce swing leg retraction in a one-legged hopper system by determining optimal inputs of shape parameters to minimize the stability measure. 
    
    \subsection{Trajectory optimization with hybrid event shaping}
    \label{sec:trajopt}
%    \begin{itemize}
%        \item Sometimes there are no shape parameters, but saltation matrix is function of time/state
%        \item In this case, stability must be designed simultaneously with trajectory optimization
%    \end{itemize}la

    Some systems do not have shape parameter terms in their saltation matrices or do not have enough to sufficiently improve stability.
    In these cases, we can change the trajectory of the system itself so that the timing, state, and input parameters of the continuous variational matrices and saltation matrices improve stability properties.
    However, it must be ensured that the dynamics, periodicity, and other constraints of the system are obeyed. 
    
    Trajectory optimization methods are a class of algorithms that aim to minimize a cost function while satisfying a set of constraints \cite{kelly_optimization}.
    For dynamical systems, these costs are generally functions of state and inputs, with constraints imposed on system dynamics and any other physical limits \cite{t-opt_methods}. 
    For specific problems, other aspects of the system may be added into the cost or constraint functions such as design parameters or minimizing time~\cite{direct_t-opt,mombaur_stable_running}. 
    Here we propose including the stability measure in the cost function to search for optimally stable trajectories. 
    Eq.~\eqref{eq:t-opt} gives the simplest form of this trajectory optimization problem, with periodicity and dynamics constraints being enforced, where dynamics constraints obey continuous dynamics in each domain and reset mappings at each hybrid event \cite{Stryk_DIRCOL}.
    Additional costs and constraints may be included such as reference tracking costs, input costs, and any physical constraints.
    We solve this problem using a direct collocation optimization with a multi-phase method to handle hybrid events \cite{kelly_optimization,t-opt_methods,betts_t-opt}.
    \begin{equation}
         \begin{array}{cl}\underset{x_{1: N},u_{1:N-1},t_{1:N-1},h_{1:M}}{\operatorname{minimize}} & \psi(\Phi(x,t,h)) \\ \text { subject to } & x_{N}=x_{1} \\ 
         & x_{i+1} = \phi(t_{i+1}-t_{i},t_{i},x_i)\end{array}
         \label{eq:t-opt}
    \end{equation}
    %This trajectory optimization method is capable of generating stable trajectories that would otherwise not be intuitively apparent. 
        
\section{Examples and Results}

    Here we demonstrate how HES can improve the stability of periodic trajectories for a variety of hybrid systems without any use of continuous-domain feedback control.
    While continuous-domain feedback could be implemented into any of these systems and should be in practice, these examples emphasize the stabilization capabilities of HES alone.
    
    The first two examples describe how previously discovered results, paddle juggling \cite{paddlejuggling} and swing leg retraction \cite{swinglegretraction}, fit into an HES framework.
    The final example demonstrates how HES can be used even without any shape parameters and how virtual hybrid events can help stabilize a complicated biped walking system.

\subsection{Paddle juggler}
    
%        \begin{itemize}
%            \item We find the stable basin as expected
%        \end{itemize}
        
    The paddle juggler system \cite{paddlejuggling}, bouncing a ball with a paddle, is known to be stabilized by impacting the ball with a paddle acceleration in a range of negative values, \eqref{eq:paddle_juggler_stable_region}, Fig.~\ref{fig:paddle_juggler}.
    The system state consists of the ball's vertical position and velocity such that $x=[x_B, \dot{x}_B]^T$. 
    This periodic hybrid system can be defined with two hybrid domains (descent and ascent) connected by two guards (impact and apex). 
    The domain $D_1$ represents the ball's aerial descent phase where $\dot{x}_B<0$ and $D_2$ represents the ball's aerial ascent phase where $\dot{x}_B>0$. 
    The guard set $g_{(1,2)} := x_B-x_P\leq0$ is defined when the ball impacts the paddle, where the paddle follows a twice differentiable trajectory $x_P(t)$. 
    The reset map $R_{(1,2)}$ is defined by a partially elastic impact law, $R_{(1,2)}([x_B,\dot{x}_B]^T)= [x_B,(1+\alpha)\dot{x}_P-\alpha\dot{x}_B]^T$,
%    \begin{equation}
%        \mathcal{R}_{(1,2)}\left(\begin{bmatrix}x_B\\\dot{x}_B \end{bmatrix}\right)=\begin{bmatrix}x_B\\(1+\alpha)\dot{x}_P-\alpha\dot{x}_B \end{bmatrix}
    %\end{equation}
    with a coefficient of restitution $\alpha$.
    The continuous dynamics in both domains follow unactuated ballistic motion: $\mathcal{F}_1=\mathcal{F}_2=[\dot{x}_B, -g]^T$, where $g$ is the acceleration due to gravity.
    
    Using these definitions, the saltation matrix \eqref{eq:salt} between domains $1$ and $2$ is constructed:
    \begin{equation}
        \Xi_{(1,2)}=\begin{bmatrix}-\alpha & 0 \\ \frac{(1+\alpha)\cdot(\ddot x_P+g)}{\dot x_P-\dot x_B} & -\alpha\end{bmatrix}
        \label{eq:Xi}
    \end{equation}
    
    Observe that $\ddot {x}_P$ appears in the saltation matrix even though it does not appear anywhere in the definition of the guards, reset maps, or vector fields of the system, making it a shape parameter that can be chosen independently of the periodic orbit.

    The guard set $g_{(2,1)} := \dot{x}_B\leq0$ is defined when the ball reaches the apex of its ballistic motion.
    Its reset map $R_{(2,1)}$ is identity and there is no change in dynamics. 
    Thus, $\Xi_{(2,1)}$ is identity and has no effect on the variations of the system.
    
    The continuous variational matrices of the system can be written exactly in closed form:
    $A_1(T)=A_2(T) = \bigl[\begin{smallmatrix}1 & T/2\\0 & 1 \end{smallmatrix}\bigr]$
    where $T$ is the total time spent in the air and also the period of the system. 
    The periodicity of the system means the ball spends an equal time, $T/2$, ascending as descending.
    
    The monodromy matrix, $\Phi$ is constructed by composing together the continuous variational matrices and saltation matrices such that $\Phi = \Xi_{(2,1)}A_2\Xi_{(1,2)}A_1$.
    
    For a given periodic bouncing trajectory, the monodromy matrix $\Phi$ is almost fully defined except for the shape parameter, $\ddot{x}_P$ in $\Xi_{(1,2)}$. 
    Given the 2-dimensional state space of this problem, the eigenvalues for any given $\ddot{x}_P$ value can be solved for explicitly.
    We can then solve exactly for where $\psi(\ddot{x}_P)<1$, giving a stable region of:
    \begin{align}
        -2g\frac{1+\alpha^2}{(1+\alpha)^2}<\ddot x_P<0
        \label{eq:paddle_juggler_stable_region}
    \end{align}
    %This stable region can also be observed in Fig. \ref{fig:paddlejugglerstability}.
    
    This is confirmed in \cite{paddlejuggling}, where the simple dynamics of the system allowed the return map to be computed explicitly without the saltation matrix.
    However, that computation is generally not possible for more complex dynamics.
    
\begin{comment}
    A specified value of $\ddot{x}_P$ is only required locally about the hybrid event.
    Away from the hybrid event, the paddle behavior can be arbitrary as long it does not induce any spurious impacts.
    A simple method to accomplish this is to fit a polynomial trajectory in a neighborhood around impact time that achieves the correct position, velocity, and acceleration at impact.
    This makes the shape parameters tractable to follow in practice.
\end{comment}
        
    \subsection{2D hopper}
        %\begin{itemize}
        %    \item 2 shape parameter inputs to be optimized
        %    \item optimization can solve for minimum stability measure
        %\end{itemize}
        
    The spring-loaded inverted pendulum (SLIP) is a popular model for dynamic legged locomotion \cite{McMahon_SLIP,HolmesLeggedLocomotion,SLIP_running}.
    This simple hopper model is effective at capturing dynamic properties of animal and robot locomotion \cite{templates_and_anchors} and has been used as a test bed for hybrid controllers \cite{clock_torque_slip}.
    
    \subsubsection{2D hopper hybrid model}
    
    \begin{figure}[!t]
        \centering
        \includegraphics[width=.9\linewidth]{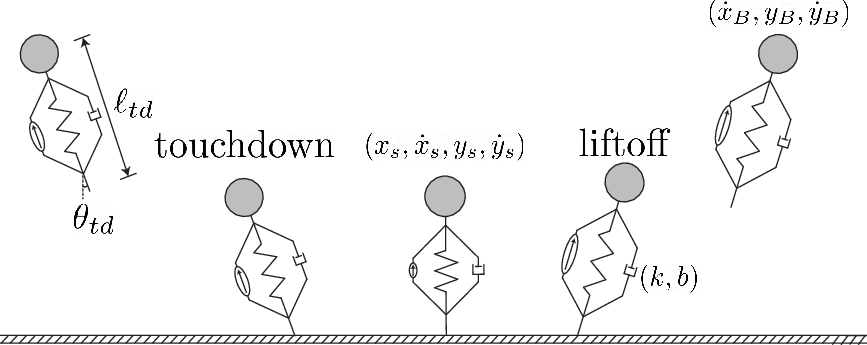}
        \caption{SLIP-like system with actuator and damper in parallel.}
        \label{fig:slip_fig}
    \end{figure}
    
    Consider a point mass body with a massless leg consisting of a spring, damper, and linear actuator all in parallel, Fig.~\ref{fig:slip_fig}. 
    This system has two domains (flight and stance) connected by two guards (touchdown and liftoff).
    The actuator is activated while in the air to preload the spring, but immediately releases at touchdown and provides no forces during stance. 
    For a periodic trajectory to occur, the actuator must preload the same amount of energy that is dissipated by the damper during stance.
    The only control authority that exists is of the leg angle in the air, which only affects the dynamics of the body at touchdown.
    
    During flight ($D_1$), the state of the hopper is represented by the horizontal velocity, vertical position, and vertical velocity of the body: $x=[\dot{x}_B,y_B,\dot{y}_B]^T$. Horizontal position is not included because it is not periodic. During stance, the body position $x_{s}$ and $y_{s}$ is defined with the toe at the origin. Horizontal position is added back into the state of the hopper such that $x=[x_{s},\dot{x}_{s},y_{s},\dot{y}_{s}]^T$.
    In flight, the dynamics of the body follow ballistic motion, while in stance there are also forces applied by the spring and damper. 
    %For simplicity, let $\ell_{s} = \sqrt{x_{s}^2+y_{s}^2}$ be the length of the leg and $\theta_{s} = \text{atan2}(x_{s},-y_{s})$ be the angle of the leg from vertical. Then let $F_s$ and $F_b$ be the forces exerted by the spring and damper, respectively such that:
    %\begin{align}
   %     F_s &=  k(\ell_0-\ell_{s})\\
   %     F_b &= -b(x_{s}\dot{x}_{s} + y_{s}\dot{y}_{s}) / \ell_{s}
    %\end{align}
%    The dynamics of each state are then:
%    \begin{align}
%        \mathcal{F}_1 & = \begin{bmatrix} 0 & \dot{y}_B & 0 \end{bmatrix}^T \\
%        \mathcal{F}_2 & = \begin{bmatrix} \dot{x}_s & %(-(F_s+F_b)*\sin(\theta+{s}))/m & \dot{y}_s & -g + (F_s+F_b)*\cos(\theta+{s})/m %\end{bmatrix}^T
%    \end{align}
    
    The touchdown guard is defined by the preload length of the leg $\ell_{td}$ and angle of the leg $\theta_{td}$ such that $g_{(1,2)} :=y_B - \ell_{td}\cos(\theta_{td})$. The liftoff guard is crossed when the force exerted by the spring-damper, $F_{sd}$, becomes zero: $g_{(2,1)} := F_{sd}$.
    There is no change in physical state of the system at the hybrid events and the reset maps only characterize the change in coordinates between domains.
\begin{comment}
    \begin{align}
        \mathcal{R}_1 & = \begin{bmatrix} -\ell_{td}\sin{\theta_{td}} & \dot{x}_B & \ell_{td}\cos{\theta_{td}} & \dot{y}_B \end{bmatrix}^T \\
        \mathcal{R}_2 & = \begin{bmatrix} \dot{x}_s & y_s & \dot{y}_s \end{bmatrix}^T
    \end{align}
\end{comment}
    %The change in coordinates also causes the saltation matrices to be non-square. $\Xi_{(1,2)}$ has dimension $4\times 3$ and $\Xi_{(2,1)}$ has dimension $3\times 4$. However, the monodromy matrix will still be square and the HES method can still operate normally.
    
    Given a set of model parameters, a trajectory from an initial condition depends only on  $\ell_{td}$ and $\theta_{td}$. 
    $\ell_{td}$ is held fixed, but $\theta_{td}$ is modulated from its nominal position $\overline{\theta}_{td}$ at time $\bar{t}_{td}$ in two ways.
    A proportional feedback term with gain K is added to stabilize the forward velocity of the system around a nominal $\overline{\dot{x}}$ and angular velocity $\dot{\theta}$ of the massless leg is also free to be chosen. $K$ and $\dot{\theta}$ are shape parameters that can be used to stabilize this system.
    \begin{align}
        \theta_{td}=\overline{\theta}_{td} + K(\dot{x}-\overline{\dot{x}}) + \dot{\theta}(t - \bar{t}_{td})
        \label{eq:leg_controller}
    \end{align}

    \subsubsection{2D hopper HES results}
    
    For a chosen initial apex height of $2.5$ with a forward velocity of $2$, $\ell_{td}$ and $\overline{\theta}_{td}$ were solved for to produce a nominal orbit,
    though the following results generalize for any choice of feasible values.
     
    With fixed shape parameters $[K,\dot{\theta} ] = [0,0]$, the system is highly unstable.
    $K$ and $\dot{\theta}$ can be optimized following \eqref{eq:shape_opt} to improve the stability of this orbit. 
    Doing so results in optimal shape parameters $[K,\dot{\theta} ] = [0.129,-0.015]$ that stabilize the trajectory, Table~\ref{table:hopper_results}. 
    Setting $\dot{\theta}=-0.015 \text{ rad/s}$ is a slow retraction rate, but there exists an interval of values $\dot{\theta}\in (-0.5892,-0.015)$ that give equivalently minimal stability measures for a fixed $K$ value.
     
     \begin{table}[!t]
     
        \caption{Stability measures of 2D hopper trajectories without and with optimized shape parameters.}
        \label{table:hopper_results}
        \centering
        \begin{tabular}{ cccc } 
        
         Shape parameters & K & $\dot{\theta}$ & Stability Measure\\
         \thickhline
         Zero & 0 & 0 & 13.756\\ 
         \hline
         Optimal (zero seed) & 0.129 & -0.015 & 0.948 \\
         \hline
         Optimal (alternate seed) & 0.129 & -0.589 & 0.948 \\
         \hline
        \end{tabular}
        
    \end{table}
    
\begin{figure}[!t]
        \centering
        \includegraphics[width=0.9\linewidth]{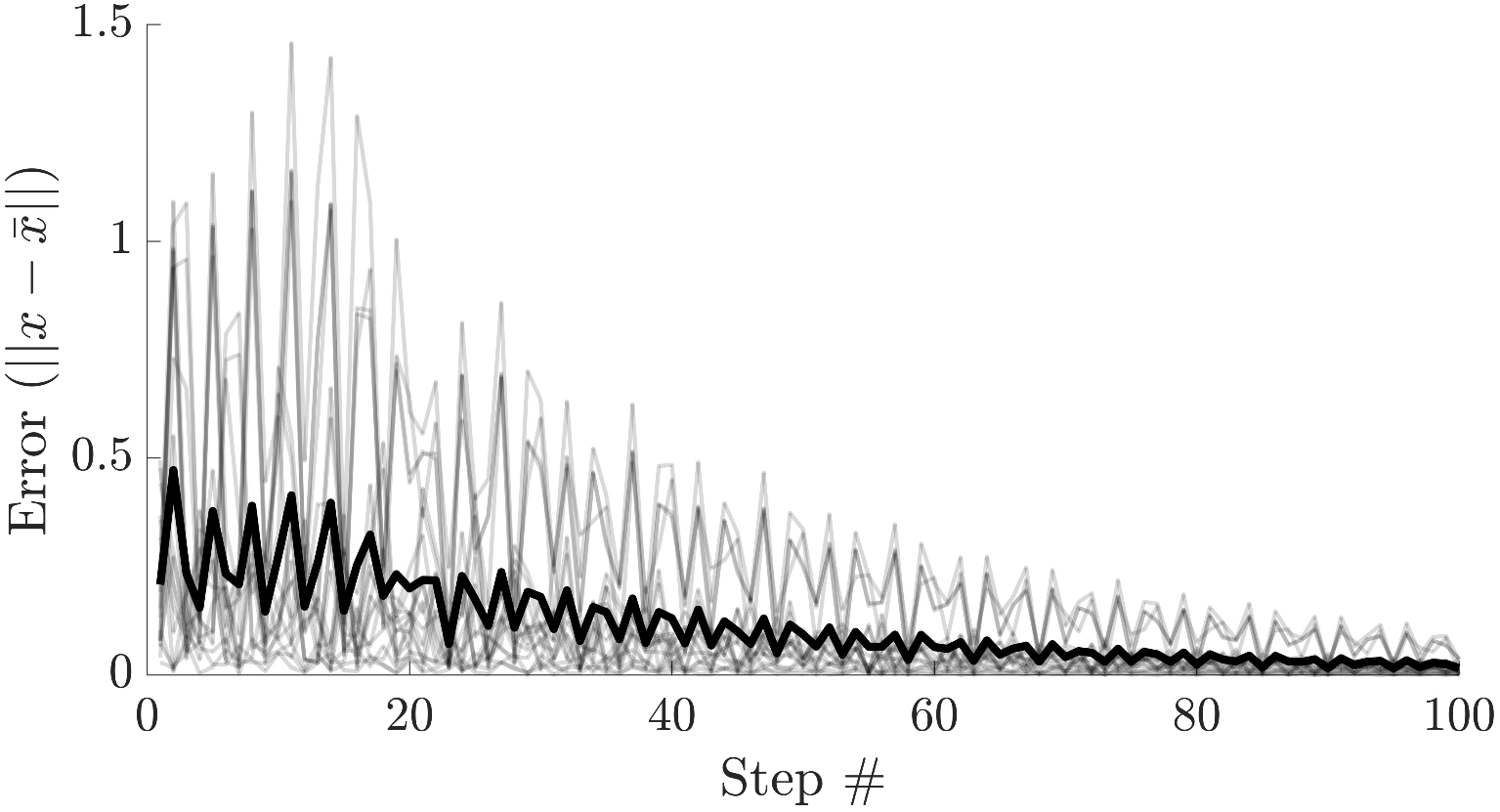}
        \caption{Error of perturbed initial states for the 2D hopper asymptotically decrease to zero. Transparent lines represent each of the 20 trials, while the bold line represents average error at each step. Convergence is neither monotonic nor very fast, but this is expected with asymptotic stability.}
        \label{fig:hopper_convergence}
    \end{figure}

    The results were confirmed in simulation by initializing 20 perturbed points in a $0.1$ radius ball around the nominal initial condition. Each of these trials was simulated for $100$ steps and the error (2-norm of the difference in perturbed state $x$ and nominal state $\bar{x}$) at apex was recorded at each step, Fig.~\ref{fig:hopper_convergence}. The zero shape parameter trajectories are not shown in the figure as every trial diverged within just 5 steps.
    
    \subsubsection{2D Hopper Discussion}
    
    %Many SLIP controllers that have been produced modulate leg orientation at impact to reject disturbances \cite{raibert,altendorfer_stability,ES-SLIP,blum_swing_leg_control}.
    The feedback term of \eqref{eq:leg_controller} is based on the Raibert stepping controller \cite{raibert}, which was utilized to great success for stabilizing early running robots.
    Other works have found that this simple controller is effective on more complex models \cite{ES-SLIP}.
    
    Another stabilizing property of legged locomotion that has been studied is swing-leg retraction \cite{swinglegretraction,optimal_swing_leg_retraction,blum_swing_leg_control}. 
    It was noted in \cite{swinglegretraction} that a 2D SLIP was able to run stably if it impacted the ground within a range negative angular leg velocities $\dot{\theta}$.
    
    The results of a negative $\dot{\theta}$ and positive $K$ agree with qualitative expectations from \cite{swinglegretraction} and \cite{raibert}.
    While a formal equivalency is yet to be proven, this is significant because the HES shape parameter optimization has no a priori knowledge that would bias its results to match these works.
    HES synthesizes two independently generated controllers and produces shape parameter values that stabilize an orbit.
    This evidence supports the potential for HES to explore other stabilizing shape parameters that are not as well studied.
    
    \subsection{Walking biped trajectory optimization}
    
    For a legged system with non-massless legs, the leg velocity shape parameter disappears as it is no longer independent of the trajectory. 
    Without shape parameters, an HES trajectory optimization can choose timing, state, and input parameters along with injecting virtual hybrid events to discover stable orbits.
    
    \subsubsection{Walking biped hybrid model}
    
\begin{figure}[!t]
        \centering
        \includegraphics[width=.9\linewidth]{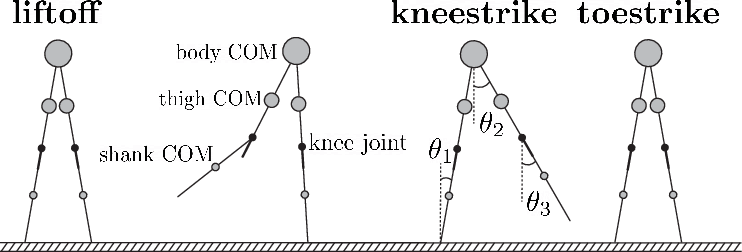}
        \caption{Biped walker system with kneestrike and toestrike hybrid events.}
        \label{fig:biped_fig}
    \end{figure}

    In this example, we consider a fully-actuated compass walker \cite{compass_walker} with knees, Fig.~\ref{fig:biped_fig}. 
    This biped model consists of two legs connected by an actuated hip joint. 
    Each leg is separated into two sections, the upper leg  (thigh) and lower leg (shank), which are connected by an actuated knee joint that has a hard stop when the thigh and shank are aligned. 
    The ankles are also actuated.

%    A point mass at the hip represents the mass of the body, while the thigh and shank of each leg are modelled with mass concentrated at their respective centers of mass.
    
    We restrict the gaits to be left-right symmetric and exclusively consist of single stance phases. 
    %This means that touchdown of the swing leg and liftoff of the stance leg will happen simultaneously, comprising just one hybrid event.
    The stance leg is locked such that its shank and thigh are aligned with each other until liftoff.
    There are 3 points of actuation at the hip, swing knee, and stance ankle.
    The state space is defined by three angles relative to vertical: stance leg, swing thigh, and swing shank, denoted $(\theta_1,\theta_2,\theta_3)$.
    
    This system has two domains. $D_1$ is the unlocked knee domain where the swing leg thigh and shank can swing freely while we enforce that $\theta_3<\theta_2$. 
    $D_2$ is the locked knee domain where the thigh and shank are constrained to be aligned with each other ($\theta_2=\theta_3$). 
    In this domain, there are only two actuators because the swing knee can no longer exert torque. The dynamics of this model are described in \cite{compass_walker}.
    
    The kneestrike guard set, between the unlocked and locked knee domains, is $g_{(1,2)} := \theta_2-\theta_3$ and
    the touchdown or toestrike guard set is $g_{(2,1)} := \theta_1+\theta_2$. 
    The reset maps at kneestrike and toestrike are also computed in \cite{compass_walker}.
    
    We add discrete changes in the inputs that induce virtual hybrid events to analyze their utility in stabilizing walking trajectories. 
    Specifically, we choose to include 5 virtual hybrid events in $D_1$ and 2 more virtual hybrid events in $D_2$, where the values of inputs between virtual hybrid events are decision variables for the optimization.
    The virtual guard functions are chosen such that $g_{v_i} := \theta_2-\theta_3+p_i$ for the first 5 virtual hybrid events and $g_{v_i} := \theta_1+\theta_2+p_i$ for the last 2 virtual hybrid events for some offset $p_i$ that is also chosen by the optimization.
    
    A direct collocation method was used with the cost consisting of the stability measure and a regularization on the input. 
    Dynamics and periodicity constraints were included along with a ground penetration constraint.
    The initial conditions of the system, given as the state after touchdown, were allowed to vary within a bounded range.
    
   \subsubsection{Walking biped HES results}
   
    In this experiment, three trajectories were compared to examine how HES can generate stable trajectories and the effect that virtual hybrid events have in further improving stability.
    A trajectory without HES was produced as a control, with its objective to minimize energy expended by using just an input regularization term in the cost.
    This minimum energy (ME) trajectory is comparable to how conventional robot locomotion trajectories are generated.
    Two HES trajectories were generated, one with virtual hybrid events (HES w/ VHE) and one without (HES w/o VHE). 

    The ME trajectory is highly unstable, while the both HES trajectories are stable with the trade off of a higher input cost.
    Specifically, HES w/ VHE has the lowest stability measure and highest energy cost, whereas HES w/o VHE was in between for both stability and cost, Table~\ref{table:trajectories}.
    
    \begin{table}[!t]
    
        \caption{Stability results for the walking biped optimization.}
        \label{table:trajectories}
        \centering
        \begin{tabular}{ ccc } 
         
         Trajectory & Stability Measure & Energy Cost\\
         \thickhline
         ME & 7.8123 & 0.985\\ 
         \hline
         HES w/o VHE  & 0.4715 & 1.337 \\
         \hline
         HES w/ VHE  & 0.3266 & 3.450 \\
        \end{tabular}
    
    \end{table}
    
\begin{figure}[!t]
        \centering
        \includegraphics[width=1\linewidth]{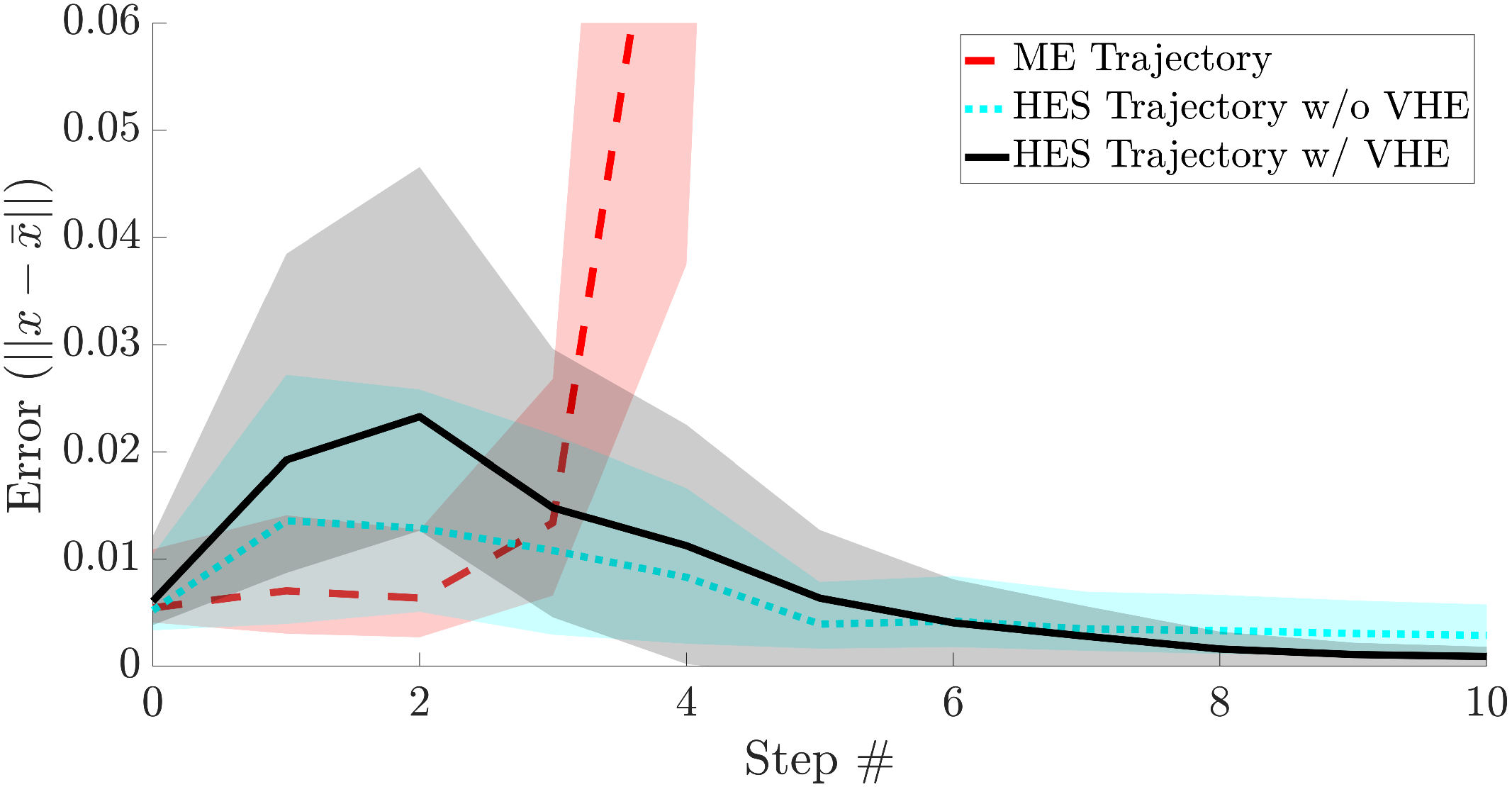}
        \caption{Error of perturbed Minimum Energy (ME), Hybrid Event Shaping without virtual hybrid events (HES w/o VHE) and HES w/ VHE trajectories over several steps. Bold lines show average error at each step and shaded regions indicate $\pm 1$ standard deviation. ME trajectories becomes highly divergent within 4 steps, while both HES trajectories appear convergent after 10 steps. The initial increase in error of the HES trajectories is allowable and is not considered by the stability measure.}
        \label{fig:bipedconvergence}
    \end{figure}

    The stability properties of the generated trajectories were confirmed through simulation.
    50 trials of each trajectory were initialized with perturbations in position and velocity between $(-0.01,0.01)$.
    Over a sequence of 10 steps, the state error at each step was tracked for each trial.
    Fig. \ref{fig:bipedconvergence} shows that after 10 steps, the HES trajectories have nearly converged back to the nominal trajectory whereas the ME trajectories diverge quickly.
    The HES w/ VHE trajectory converges to a smaller error after 10 steps compared to the HES w/o VHE trajectory, which supports the findings of the stability measure.
    %Both HES trajectories experience an increase in average error over the first few steps, with the error of the HES with VHE trajectory increasing more. 
    %These increases are allowable under the stability criterion and are not captured by the stability measure.
    %Notice that for the first few steps, the average divergence increases due to the non-contractiveness of the system.
    
    %\begin{figure}[!t]
        %\centering
        %\includegraphics[width=1\linewidth]{Biped_step_to_step.jpg}
        %\caption{Perturbed states over several steps}
        %\label{fig:bipeddivergence}
    %\end{figure}
    
    \section{Conclusion}
    
    While the idea of hybrid event control is not novel, hybrid event shaping provides a generalized method to analyze the stability of hybrid orbits and select hybrid event parameters to optimize stability. 
    HES unifies results of previous simple hybrid event controllers while also being compatible with trajectory optimization techniques to produce stable trajectories for complex systems.
    HES computes the derivative of the stability measure, improving computational efficiency compared to previous stability optimization methods.
    Compared to previous work, HES does not rely on human observation and tuning to design stabilizing hybrid event parameters.
    
    %Some work has been done to numerically search for optimally stable shape parameters without the saltation matrix\cite{optimal_swing_leg_retraction}, but for higher dimensional systems, these numerical methods become untenable, whereas HES can still succeed due to derivative computations. 
    %Additionally, these numerical methods can not identify all the shape variables of a system.
    %The HES method is able to define the effect of each shape parameter and uses the shape parameter optimization to determine the optimal values of these inputs.
    
    In this work, there was no use of continuous-domain feedback that is commonly utilized in hybrid systems control. We believe that hybrid event shaping is one aspect that can be used in conjunction with continuous-domain feedback to improve the success rate of robots performing dynamic behaviors in real-world settings. 
    This is not be prohibitively complex because saltation matrices are not affected by feedback control laws in the continuous domains. 
    In the future, we aim to synthesize HES methods with continuous-domain feedback control to produce even more stable closed-loop trajectories.
    We also found that virtual hybrid events may have utility in stabilizing hybrid orbits, but design factors such as how many virtual hybrid events to insert and when to insert them merit further investigation.
    
    %\begin{itemize}
        %\item HES is pretty versatile and captures many interesting aspects developed at hybrid events
        %\item Continuous-domain feedback not used at all in this work; can and probably should be for even better performance. But not strictly neccessary 
        %\item Robotocists should spend more time thinking about what exactly happens at hybrid events and how to use them to our advantage
    %\end{itemize}
    
    %\section{Future Work}
    
    %\begin{itemize}
       % \item approximating shape parameters
        %\item HES with continuous-domain feedback: robustness?
        %\item more efficient algorithm
        %\item distributed control stuff?
        %\item controlability?
        %\item mode sequence
    %\end{itemize}
\addtolength{\textheight}{-4.0cm}   % This command serves to balance the column lengths
                                  
\bibliographystyle{IEEEtran}
\bibliography{ref}

\begin{thebibliography}{10}
\providecommand{\url}[1]{#1}
\csname url@samestyle\endcsname
\providecommand{\newblock}{\relax}
\providecommand{\bibinfo}[2]{#2}
\providecommand{\BIBentrySTDinterwordspacing}{\spaceskip=0pt\relax}
\providecommand{\BIBentryALTinterwordstretchfactor}{4}
\providecommand{\BIBentryALTinterwordspacing}{\spaceskip=\fontdimen2\font plus
\BIBentryALTinterwordstretchfactor\fontdimen3\font minus
  \fontdimen4\font\relax}
\providecommand{\BIBforeignlanguage}[2]{{%
\expandafter\ifx\csname l@#1\endcsname\relax
\typeout{** WARNING: IEEEtran.bst: No hyphenation pattern has been}%
\typeout{** loaded for the language `#1'. Using the pattern for}%
\typeout{** the default language instead.}%
\else
\language=\csname l@#1\endcsname
\fi
#2}}
\providecommand{\BIBdecl}{\relax}
\BIBdecl

\bibitem{Altin_hybrid_MPC}
B.~{Altın}, P.~{Ojaghi}, and R.~G. {Sanfelice}, ``A model predictive control
  framework for hybrid dynamical systems,'' \emph{IFAC-PapersOnLine}, vol.~51,
  no.~20, pp. 128--133, 2018.

\bibitem{kong2021ilqr}
N.~J. Kong, G.~Council, and A.~M. Johnson, ``{iLQR} for piecewise-smooth hybrid
  dynamical systems,'' in \emph{IEEE Conference on Decision and Control},
  December 2021, to appear.

\bibitem{invariant_impact_control}
W.~Yang and M.~Posa, ``Impact invariant control with applications to bipedal
  locomotion,'' \emph{arXiv preprint arXiv:2103.06907}, 2021.

\bibitem{Burden_dimension_reduction}
S.~{Burden}, S.~{Revzen}, and S.~{Sastry}, ``Dimension reduction near periodic
  orbits of hybrid systems,'' in \emph{IEEE Conference on Decision and
  Control}, September 2011.

\bibitem{dbhop}
G.~{Council}, S.~{Yang}, and S.~{Revzen}, ``Deadbeat control with (almost) no
  sensing in a hybrid model of legged locomotion,'' in \emph{International
  Conference on Advanced Mechatronic Systems}, 2014, pp. 475--480.

\bibitem{paddlejuggling}
D.~{Sternad}, M.~{Duarte}, H.~{Katsumata}, and S.~{Schaal}, ``Bouncing a ball:
  Tuning into dynamic stability,'' \emph{Journal of Experimental Psychology:
  Human Perception and Performance}, vol.~27, pp. 1163--84, 11 2001.

\bibitem{swinglegretraction}
A.~{Seyfarth}, H.~{Geyer}, and H.~{Herr}, ``Swing-leg retraction: a simple
  control model for stable running,'' \emph{Journal of Experimental Biology},
  vol. 206, no.~15, pp. 2547--2555, 2003.

\bibitem{ankarali2010control}
M.~M. Ankaral{\i}, U.~Saranl{\i}, and A.~Saranl{\i}, ``Control of underactuated
  planar hexapedal pronking through a dynamically embedded slip monopod,'' in
  \emph{IEEE International Conference on Robotics and Automation}, 2010, pp.
  4721--4727.

\bibitem{Green_2020}
K.~{Green}, R.~{Hatton}, and J.~{Hurst}, ``Planning for the unexpected:
  Explicitly optimizing motions for ground uncertainty in running,'' in
  \emph{IEEE International Conference on Robotics and Automation}, May 2020.

\bibitem{vanderschaft_hybrid_systems}
A.~{van der Schaft} and J.~{Schumacher}, ``Complementarity modeling of hybrid
  systems,'' \emph{IEEE Transactions on Automatic Control}, vol.~43, no.~4,
  1998.

\bibitem{lygeros_hybrid_automata}
J.~{Lygeros}, K.~H. {Johansson}, S.~{Simic} \emph{et~al.}, ``Dynamical
  properties of hybrid automata,'' \emph{IEEE Transactions on Automatic
  Control}, 2003.

\bibitem{johnson_hybrid_systems}
A.~M. {Johnson}, S.~A. {Burden}, and D.~E. {Koditschek}, ``A hybrid systems
  model for simple manipulation and self-manipulation systems,'' \emph{The
  International Journal of Robotics Research}, vol.~35, no.~11, pp. 1354--1392,
  2016.

\bibitem{Joyce2012}
D.~Joyce, ``{On manifolds with corners},'' in \emph{Advances in Geometric
  Analysis}, ser. Advanced Lectures in Mathematics.\hskip 1em plus 0.5em minus
  0.4em\relax International Press of Boston, Inc., 2012, vol.~21, pp. 225--258.

\bibitem{sliding}
M.~{Jeffrey}, ``Dynamics at a switching intersection: Hierarchy, isonomy, and
  multiple sliding,'' \emph{SIAM Journal on Applied Dynamical Systems},
  vol.~13, pp. 1082--1105, 07 2014.

\bibitem{simic_towards_2000}
S.~N. Simić, K.~H. Johansson, S.~Sastry, and J.~Lygeros, ``Towards a
  {Geometric} {Theory} of {Hybrid} {Systems},'' in \emph{Hybrid {Systems}:
  {Computation} and {Control}}, ser. Lecture {Notes} in {Computer} {Science},
  N.~Lynch and B.~H. Krogh, Eds.\hskip 1em plus 0.5em minus 0.4em\relax Berlin,
  Heidelberg: Springer, 2000, pp. 421--436.

\bibitem{zeno}
J.~{Zhang}, K.~H. {Johansson}, J.~{Lygeros}, and S.~{Sastry}, ``Dynamical
  systems revisited: Hybrid systems with zeno executions,'' in \emph{Hybrid
  Systems: Computation and Control}.\hskip 1em plus 0.5em minus 0.4em\relax
  Springer Berlin Heidelberg, 2000.

\bibitem{aizerman_hybrid_stability}
\BIBentryALTinterwordspacing
M.~{Aizerman}. and F.~{Gantmacher}, ``{Determination of Stability by Linear
  Approximation of a Periodic Solution of a System of Differential Equations
  With Discontinuous Right-hand Sides},'' \emph{The Quarterly Journal of
  Mechanics and Applied Mathematics}, vol.~11, no.~4, pp. 385--398, 01 1958.
  [Online]. Available: \url{https://doi.org/10.1093/qjmam/11.4.385}
\BIBentrySTDinterwordspacing

\bibitem{contraction}
S.~A. Burden, T.~Libby, and S.~D. Coogan, ``On contraction analysis for hybrid
  systems,'' \emph{arXiv preprint arXiv:1811.03956}, 2018.

\bibitem{Leine}
R.~{Leine} and H.~{Nijmeijer}, \emph{Dynamics and Bifurcations of Non-Smooth
  Mechanical Systems}.\hskip 1em plus 0.5em minus 0.4em\relax Springer-Verlag
  Berlin Heidelberg, 2004.

\bibitem{hirsch2012differential}
M.~{Hirsch}, S.~{Smale}, and R.~{Devaney}, \emph{Differential Equations,
  Dynamical Systems, and an Introduction to Chaos}.\hskip 1em plus 0.5em minus
  0.4em\relax Academic Press, 2012.

\bibitem{Muller_lyapunov_exponents}
P.~Müller, ``Calculation of lyapunov exponents for dynamic systems with
  discontinuities,'' \emph{Chaos, Solitons \& Fractals}, vol.~5, no.~9, pp.
  1671--1681, 1995.

\bibitem{monodromy}
X.~{Wang} and J.~K. {Hale}, ``On monodromy matrix computation,'' \emph{Computer
  Methods in Applied Mechanics and Engineering}, vol. 190, no.~18, 2001.

\bibitem{monodromy_matrix}
H.~{Asahara} and T.~{Kousaka}, ``Stability analysis using monodromy matrix for
  impacting systems,'' \emph{IEICE Transactions of Fundamentals of Electronics,
  Communications, and Computer Science}, vol. E101-A, no.~6, 2018.

\bibitem{optimal_swing_leg_retraction}
J.~G.~D. {Karssen}, M.~{Haberland}, M.~{Wisse}, and S.~{Kim}, ``The optimal
  swing-leg retraction rate for running,'' in \emph{IEEE International
  Conference on Robotics and Automation}, 2011, pp. 4000--4006.

\bibitem{IntrotoNumericalAnalysis}
E.~{Tyrtyshnikov}, \emph{A Brief Introduction to Numerical Analysis}.\hskip 1em
  plus 0.5em minus 0.4em\relax Birkhäuser Basel, 1997.

\bibitem{eig_derivatives}
P.~{Lancaster}, ``On eigenvalues of matrices dependent on a parameter,''
  \emph{Numerische Mathematik}, 1964.

\bibitem{Khalil}
H.~{Khalil}, \emph{Nonlinear Systems}.\hskip 1em plus 0.5em minus 0.4em\relax
  Pearson, 2002.

\bibitem{num_opt}
J.~{Nocedal} and S.~{Wright}, \emph{Numerical Optimization}.\hskip 1em plus
  0.5em minus 0.4em\relax Springer, 2006.

\bibitem{kelly_optimization}
M.~{Kelly}, ``An introduction to trajectory optimization: How to do your own
  direct collocation,'' \emph{SIAM Review}, vol.~59, no.~4, pp. 849--904, 2017.

\bibitem{t-opt_methods}
O.~{Von Stryk} and R.~{Bulirsch}, ``Direct and indirect methods for trajectory
  optimization,'' \emph{Annals of Operations Research}, vol.~37, pp. 357--373,
  12 1992.

\bibitem{direct_t-opt}
C.~R. {Hargraves} and S.~W. {Paris}, ``Direct trajectory optimization using
  nonlinear programming and collocation,'' \emph{Journal of Guidance, Control,
  and Dynamics}, vol.~10, no.~4, pp. 338--342, 1987.

\bibitem{mombaur_stable_running}
K.~{Mombaur}, R.~{Longman}, H.~{Bock}, and J.~{Schlöder}, ``Open-loop stable
  running,'' \emph{Robotica}, vol.~23, pp. 21--33, 01 2005.

\bibitem{Stryk_DIRCOL}
O.~V. {Stryk}, ``User's guide for {DIRCOL} (version 2.1): a direct collacation
  method for the numerical solution of optimal control problems,'' in
  \emph{Technische Universitat Darmstadt}, November 1999.

\bibitem{betts_t-opt}
J.~T. {Betts}, \emph{Practical Methods for Optimal Control and Estimation Using
  Nonlinear Programming, Second Edition}, 2nd~ed.\hskip 1em plus 0.5em minus
  0.4em\relax SIAM, 2010.

\bibitem{McMahon_SLIP}
T.~A. {McMahon} and G.~C. {Cheng}, ``The mechanics of running: How does
  stiffness couple with speed?'' \emph{Journal of Biomechanics}, vol.~23, pp.
  65--78, 1990.

\bibitem{HolmesLeggedLocomotion}
P.~{Holmes}, R.~J. {Full}, D.~E. {Koditschek}, and J.~{Guckenheimer}, ``The
  dynamics of legged locomotion: Models, analyses, and challenges,'' \emph{SIAM
  Review}, vol.~48, no.~2, pp. 207--304, 2006.

\bibitem{SLIP_running}
A.~{Seyfarth}, H.~{Geyer}, M.~{Günther}, and R.~{Blickhan}, ``A movement
  criterion for running,'' \emph{Journal of Biomechanics}, vol.~35, no.~5, pp.
  649--655, 2002.

\bibitem{templates_and_anchors}
R.~J. {Full} and D.~E. {Koditschek}, ``{Templates and anchors: neuromechanical
  hypotheses of legged locomotion on land},'' \emph{Journal of Experimental
  Biology}, vol. 202, no.~23, pp. 3325--3332, 12 1999.

\bibitem{clock_torque_slip}
J.~{Seipel} and P.~{Holmes}, ``{A simple model for clock-actuated legged
  locomotion},'' \emph{Regular and Chaotic Dynamics}, vol.~12, no.~5, Oct.
  2007.

\bibitem{raibert}
M.~{Raibert}, H.~{Brown}, and M.~{Chepponis}, ``Experiments in balance with a
  {3D} one-legged hopping machine,'' \emph{The International Journal of
  Robotics Research}, vol.~3, no.~2, pp. 75--92, 1984.

\bibitem{ES-SLIP}
I.~{Poulakakis} and J.~W. {Grizzle}, ``The spring loaded inverted pendulum as
  the hybrid zero dynamics of an asymmetric hopper,'' \emph{IEEE Transactions
  on Automatic Control}, vol.~54, no.~8, pp. 1779--1793, 2009.

\bibitem{blum_swing_leg_control}
Y.~Blum, S.~Lipfert, J.~Rummel, and A.~Seyfarth, ``Swing leg control in human
  running,'' \emph{Bioinspiration \& Biomimetics}, vol.~5, June 2010.

\bibitem{compass_walker}
V.~Chen, ``Passive dynamic walking with knees: A point foot model,'' Master's
  thesis, Massachusetts Institute of Technology, 2007.

\end{thebibliography}

\end{document}